\documentclass[10pt,twocolumn,letterpaper]{article}

\usepackage{cvpr}
\usepackage{times}
\usepackage{epsfig}
\usepackage{graphicx}
\usepackage{amsmath}
\usepackage{amssymb}

\usepackage[breaklinks=true,bookmarks=false]{hyperref}

\cvprfinalcopy 

\def\cvprPaperID{****} 

\begin{document}

\title{FaceLiveNet+: A Holistic Networks For Face Authentication Based On Dynamic Multi-task Convolutional Neural Networks}

\author{Zuheng Ming\\
University of La Rochelle\\
\and
Junshi Xia\\
RIKEN\\
\and
Muhammad Muzzamil Luqman\\
University of La Rochelle\\
\and
Jean-Christophe Burie\\
University of La Rochelle\\
\and
Kaixing Zhao\\
University of Toulouse\\
}

\maketitle

\begin{abstract}
This paper proposes a holistic multi-task Convolutional Neural Networks (CNNs) 
with the dynamic weights of the tasks,
%
namely FaceLiveNet+, for face authentication.  FaceLiveNet+ can employ face verification and facial expression recognition as a solution of liveness control simultaneously. Comparing to the single-task learning,  the proposed multi-task learning can better capture the feature representation for all of the tasks. The experimental results show the superiority of the multi-task learning to the single-task learning for both the face verification task and facial expression recognition task. Rather using a conventional multi-task learning with fixed weights for the tasks, this work proposes a so called dynamic-weight-unit to automatically learn the weights of the tasks. The experiments have shown the effectiveness of the dynamic weights for training the networks. Finally, the holistic evaluation for face authentication based on the proposed protocol has shown the feasibility to apply the FaceLiveNet+ for face authentication. 
\end{abstract}

\section{Introduction}
Benefiting from the progress of the representing learning with the deep CNNs, the face-related recognition problems have made remarkable progress recently ~\cite{taigman2014deepface, parkhi2015deep, schroff2015facenet, liu2017sphereface}. These works have achieved or beyond the human-level performance on the benchmarks LFW\cite{huang2007labeled}, YTF\cite{wolf2011face} or more challenged MegaFace~\cite{kemelmacher2016megaface}, which greatly boost the application of the face-related biometric authentication with the advantages of less human-invasion and easy to use.

\begin{figure}[t]
\begin{center}
   \includegraphics[width=0.9\linewidth]{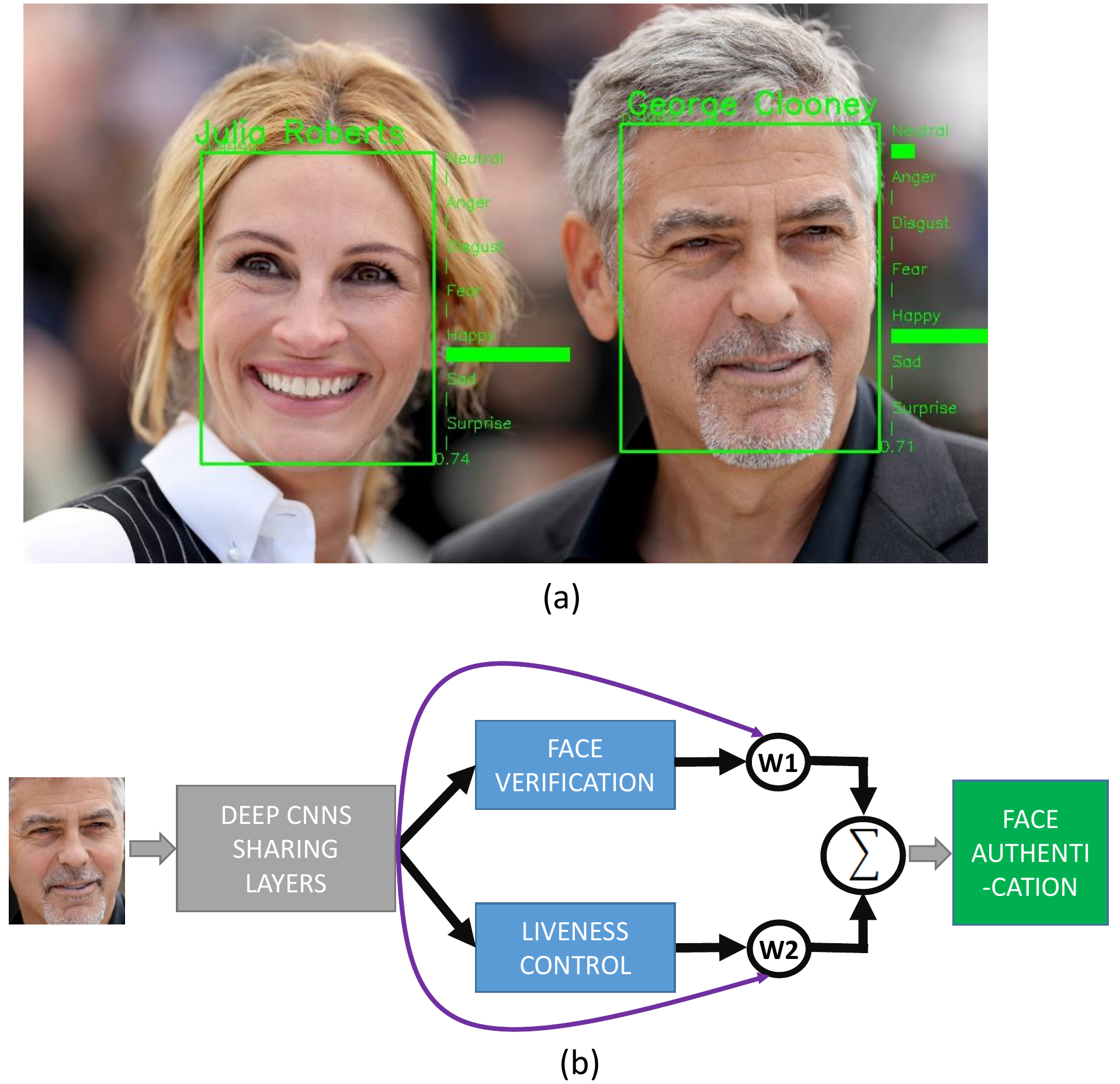}
\end{center}
   \caption{(a) The proposed multi-task FaceLiveNet+ enable to perform face verification and facial expression recognition as liveness control simultaneously for face authentication. (b) The proposed multi-task framework can jointly learn the two tasks with the dynamic weights learned automatically in the training processing.}
\label{framework}
\end{figure} 

Face verification is an indispensable part of a face authentication system, however the liveness control which aims to detect the real presence of the user before the camera is not always included in the system or is included as an independent stage in a sequential protocol after the face verification stage ~\cite{wen2015face,pan2007eyeblink}. The stand-alone framework in a chain protocol is vulnerable in the spoof-attack with a photo or video, in which the fraudster can attack the system with the photo of the real user for passing face verification and then pass the liveness control by presenting himself physically before the camera to satisfy the liveness detection, e.g. the detection of moving the head or blinking the eye etc.  Thus a multi-task networks which can conduct the tasks of face verification and the liveness detection simultaneously is essential for face authentication (see \figurename~\ref{framework}). 

Many methods have been proposed for the liveness detection. The stability, the complexity and the requirement of these methods vary greatly. The emerging 3D structure light based solution can well protect the system from the 2D photo or screen attack by detecting the 3D depth information, however it cannot be easily realized on most of the current devices by requiring for example the assistance of infrared sensors.~\cite{pribanic20163d}. The motion-based detection methods such as the detection of the eye blinking~\cite{pan2007eyeblink}, the head pose~\cite{frischholz2003avoiding}, and the face motion~\cite{bharadwaj2013computationally} can be easily implemented in a low-cost way, but the simple motion-based detection can be easily spoofed with a video downloaded from the social networks. Other methods leverage the existing datasets to train a model to analyze the difference of the image texture in terms of specular reflection, image blurriness, image chromaticity etc, to distinguish the fraudulent and real images~\cite{zhang2012face, chingovska2012effectiveness}. Nevertheless, the limit of the generalization capacity has been shown in the cross evaluation between the datasets~\cite{wen2015face}. As a trade-off between the capacity, the facility and the stability, the facial expression recognition methods based on the challenge-response mechanism has been proposed for the liveness control, in which the system would detect whether the user in front of the camera can play a specific expression asked by the system. In this work, the required expressions are limited among the six posed universal expressions, i.e. Neutral, Anger, Disgust, Fear, Happy, Sad and Surprise.    

Before this work, the multi-task learning of face verification and face verification has not been extensively studied. Unlike the variations of the pose of the face, the facial expression introduce deformations of the face which results in the difficulty for face recognition. With the merit of eavesdropping, the multi-task learning can capture a representation of features being difficult learned by one task but can easily learned by another task~\cite{ruder2017overview}. In this work, we also aim to leverage the multi-task learning to boost the performances of the two tasks in comparison with a single-task configuration. According to the architecture to perform the multi-task in deep neural networks, the multi-task learning is typically divided into two categories: hard parameters sharing and soft parameters sharing. Hard parameter sharing, which greatly reduces the overfitting risk~\cite{baxter1997bayesian}, is the most commonly used approach of multi-task learning in neural networks~\cite{caruna1993multitask}. We also adopt the hard parameters sharing in this work in which the tasks share the common hidden layers in the front part of the network and keep the task-specific layers as branches of the end of the network (see \figurename~\ref{fig:architeture}).  How to set the weights of the tasks in the multi-task learning is an important issue. The weights determine the importance of the different tasks in the holistic networks. Many works experimentally set the weights of the tasks or simply set the equal values for all tasks.  Hyperface~\cite{ranjan2017hyperface} manually set the weights of the  tasks such as the face detection, landmarks localization, pose estimation and gender recognition according to their importance in the overall loss. In ~\cite{tian2015pedestrian}, the authors obtain the optimal weights by a greedy search for pedestrian detection tasks with the different attributes. In ~\cite{chen2017multi}, the authors assign equal weights to the ranking task and the binary classification task for the person re-identification. 
All these methods set the fixed weights to optimize the tasks, however the importance of the tasks are probably varied during the training processing. The easy task can be trained firstly  with a lager weight and then, the weight of the hard task tends to increase to further optimizing the global performance.
Inspired by~\cite{misra2016cross, yin2018multi}, we propose to set the weights of the tasks as the parameters of the neural networks which can be learned during the training process so that the importance of the tasks can be measured dynamically during the training. In our proposed network we also keep a branch for each task. 

Our main contributions are summarized as follows.
\begin{itemize}
	\item We have proposed FaceLiveNet+, a multi-task deep CNNs-based networks with dynamically learned weights which can conduct simultaneously face verification and facial expression recognition for face authentication. 
    \item We have demonstrated that, for both face verification and facial expression recognition tasks, FaceLiveNet+ can achieve the state-of-the-art or better performance on the datasets LFW\cite{huang2007labeled} and YTF\cite{wolf2011face}, CK+~\cite{lucey2010extended}, OuluCASIA~\cite{zhao2011facial}. In comparison with the single-task configuration or the multi-task deep networkswith fixed weights, the performance are boosted.
    \item We have proved the effectiveness of the dynamic weights of the tasks for training the multi-task networks. 
\end{itemize}

The remainder of this paper is organized as follows: Section II briefly reviews the related works; Section III describes the architecture of the dynamic multi-task network. Section IV presents the training approach following by Section V where the experimental results are analyzed. Finally, in Section VI, we draw the conclusions and present the future works.

\section{Related works}
\subsection{Face recognition}
Face recognition is one of the most studied research area in the past decades. The conventional methods for face recognition before deep representation learning have mainly focused on the feature engineering to design the best handcrafted features to classify the faces, such as the local descriptors LBP, Gabor-LBP, HOG, SIFT~\cite{ahonen2006face, deniz2011face, bicego2006use}. The Fisher Vector~\cite{simonyan2013fisher} proposed to fuse the different features into an overall face descriptor. Since the coming of the  representation learning with the deep CNNs, face recognition has made a series breakthrough via the the deep CNNs. DeepFace~\cite{taigman2014deepface} introduced the siamese networks for face verification and uses a 3D model to align the faces. It has achieved 97.35\% on the LFW and 91.4\% on the YTF. DeepID~\cite{sun2015deeply} proposed 3 series of face recognition models based on deep CNNs. The significant feature of the DeepID series was to use more than 200 CNNs for face verification and gained a better performance (99.15\% on LFW). FaceNet~\cite{schroff2015facenet} is one of the most cited work in the recent years for face recognition. FaceNet proposed to use the triplet loss to redistribute the embedding feature space which can be used to the face-related classification problem. Benefiting from the enormous private training data, Facenet has achieved the state-of-art on LFW (99.63\%) and YTF (95.12\%). However, the computation cost for calculating the triplet loss has promoted the works~\cite{wen2016discriminative, ming2017simple} to simplify the algorithm. VGGFace~\cite{simonyan2014very} continued to implement the triplet loss on the VGG networks with 16 layers. VGGFace is recently one of the most used pretrained model in the face related problem.  SphereFace~\cite{liu2017sphereface} proposed the angular softmax to learn angularly discriminative features and has achieved the state-of-the-art performance on dataset MegaFace.

\subsection{Liveness control}
The liveness control has mainly four categories methods: 1) motion detection methods, which detect the motion of the subject including  the eye-blinking~\cite{pan2007eyeblink}, the head movement~\cite{frischholz2003avoiding} or the face motion~\cite{bharadwaj2013computationally} to effect the liveness detection. However the simple motion-based detection method can be either easily fooled with the crude photo-attack ~\cite{boulkenafet2017oulu}. 2) texture analysis that can detect the  texture  differences such as  the  specular  reflection,  image  blurriness,  image  chromaticity between the images of photo/video  spoof and the real individual~\cite{wen2015face}. However the generalization capacity of these methods are their weakness. 3) interactive methods based on Challenge-Response mechanism can achieve a better generalization capacity by interacting with the users to response the requirement of the system~\cite{frischholz2003avoiding}. 4) 3D methods to reconstruct the 3D face model by the obtained 3D cloud points. The 3D methods can easily detect the flat surface such as the screen or photo while the assistance of sensors such as the structure light projector~\cite{pribanic20163d} or the Time-of-Flight (ToF) sensor~\cite{cui20103d}, in order to capture the cloud points.   

\subsection{Facial expression recognition}

As face recognition, the deep CNNs has greatly advanced the state-of-the-art performance of facial expression recognition. Comparing to the previous works based on the Support Vector Machine (SVM) with the hand-crafted features~\cite{feng2007facial, klaser2008spatio, levi2015emotion}, the deep CNNs-based methods have significantly improved the performance~\cite{jung2015joint,zhao2016peak, mollahosseini2016going}. Unlike the face recognition issue, the datasets for facial expression recognition have much less number of the images such as CK+ and OuluCASIA. To the best of our knowledge, even so far the largest dataset FER2013~\cite{goodfellow2013challenges} of natural expression collected from the internet has only about 20 thousands images. It is difficult to train a deep CNNs based model from scratch for facial expression recognition with such small datasets. Thus the facial expression model based on deep CNNs has normally transferred from the face recognition model pretrained on the relative larger datasets such as CASIA-Webface~\cite{yi2014learning}, MSCeleb-1M~\cite{guo2016ms}, etc. 

\subsection{Multi-task learning}
Multi-task learning not only helps to learn more than one task in a single network but also can improve upon your main task with an auxiliary task~\cite{ruder2017overview}. In this work, we focus on the multi-task learning in the context of the deep CNNs. Hyperface~\cite{ranjan2017hyperface} proposed a multi-task learning algorithm for face detection, landmarks localization, pose estimation and gender recognition using deep CNNs. The tasks have been set the different weights according to the importance of the task. The experiments results shown that the sharing of the fuse features boost the performance for all four tasks comparing to the single-task learning. ~\cite{chen2017multi} integrated the classification task and the ranking task in a multi-task networks for person re-identification problem. By jointly optimizing the two tasks simultaneously, the performance has been significant improved. Each task has been set with a equal weight. Zhang et al.~\cite{zhang2014facial} used the multi-task based CNN for employing the detection of the facial landmark, the estimation of the discrete head yaw, the recognition of gender and the detection of smile and glass. Unlike HyperFace, the predictions for all these tasks were pooled from the same feature space. Yim et al. ~\cite{yim2015rotating} adopted equal weights for the tasks of face recognition and face frontalization. Tian et al.~\cite{tian2015pedestrian} fix the weight for the main task to 1, and obtain the weights of all side tasks via a greedy search within 0 and 1. However, the greedy search space is so large that the computation is time consuming. Xi et al.~\cite{yin2018multi} proposed a multi-task model for face pose-invariant recognition with an automatic learning of the weights for each task. Misra et al.~\cite{misra2016cross} proposed to use the cross-stitch units to allow the model to determine in what way the
task-specific network leverage the knowledge of the other task by learning a linear combination of the output of the previous layers.

\section{Architecture}
\begin{figure}[t]
\begin{center}
   \includegraphics[width=1.9\linewidth, angle =90]{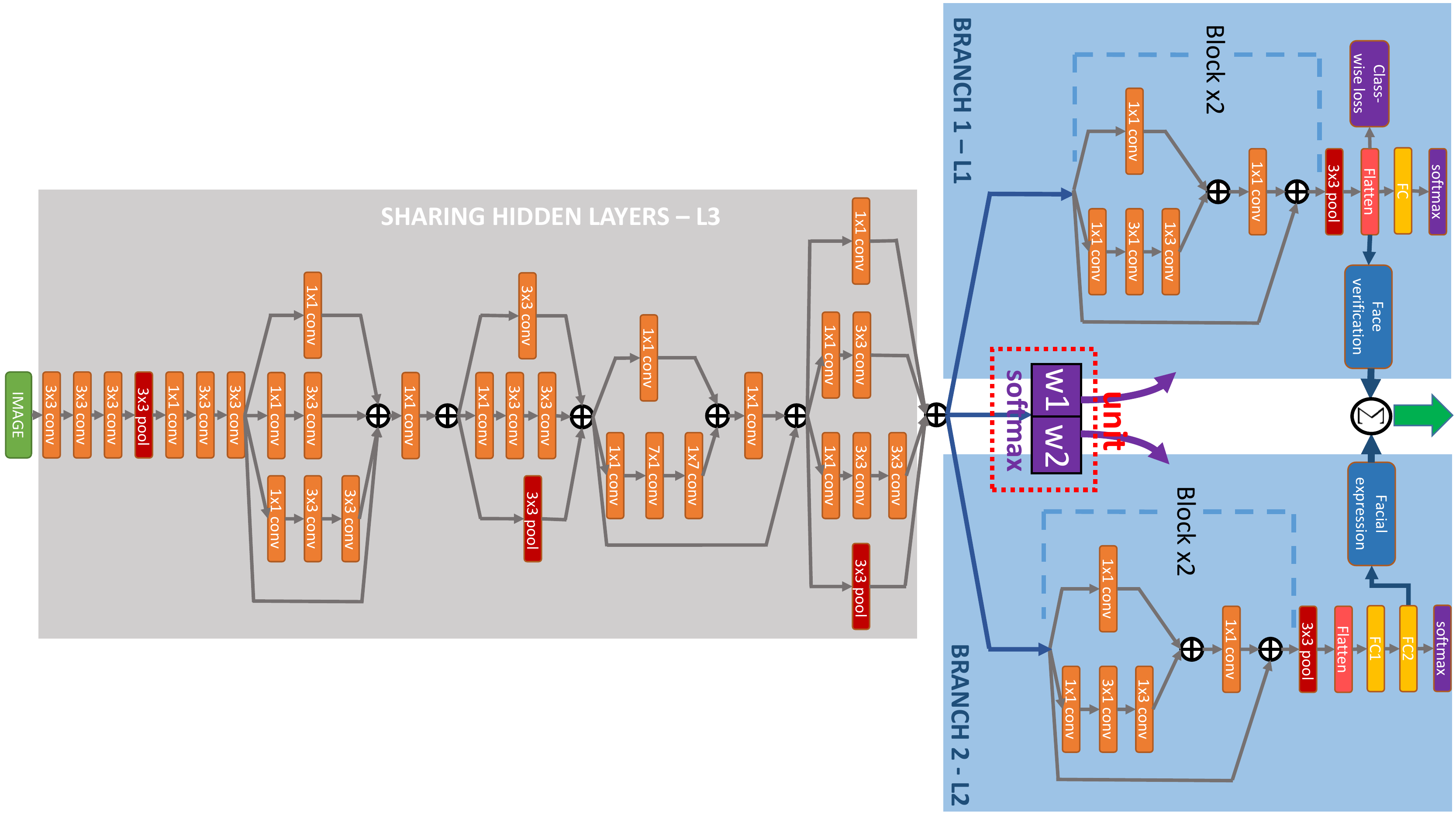}
\end{center}
   \caption{The architecture of the proposed multi-task FaceLiveNet+ which can jointly learn the two tasks of face verification and facial expression recognition. The weights of the tasks can be automatically learned by the dynamic-weights-unit. In the application phase, FaceLiveNet+ can perform face verification and facial expression recognition simultaneously.}
\label{fig:architeture}
\label{fig:onecol}
\end{figure}

The proposed dynamic multi-task network (see ~\figurename~\ref{fig:architeture} is based on the hard parameter sharing structure in which the sharing hidden layers are shared between all tasks~\cite{ruder2017overview}. The task-specific layers consisting of two branches are respectively dedicated to face verification and facial expression recognition. The face authentication can be conducted by the fusion of the results of face verification and facial expression recognition. The two branches have almost identical structures to facilitate the transfer learning of facial expression recognition from the pretrained face recognition task. Specifically, the BRANCH 1 extracts the embedded features of bottleneck layer for face verification and the BRANCH 2 has one more fully connected layer than BRANCH 1 serving to calculate the probabilities of the facial expressions. The deep CNNs based on the Inception-ResNet have 13 million parameters of about 30 hidden layers. The size of the parameters is fewer than other popular deep CNNs such as VGGFace or FaceNet.

\textbf{Dynamic-weight-unit} The unit between the two branches is used to generate the dynamic weights of each task. This unit connected to the last layer of the sharing hidden layers can be treated as a new branch with only one fully connected layer. The size of the unit is equal to the number of weights, e.g. the size is 2 in this work. In order to normalize the weights to [0,1] with a sum of 1, i.e. $w1+w2=1$, the softmax function has been applied on this layer. Consequently, the weights are actually the outputs of this softmax layer. The parameters of this softmax layer are optimized during the training of the networks with the weighted loss. Instead of using fixed weights, the weights of the tasks in this framework can be automatically learned with a dynamic variation.     

\section{Training approach}
~\subsection{Dynamic Multi-task CNNs}
The different loss functions are applied to train the different tasks in the proposed multi-task CNNs, and the overall loss function is the sum of the weighted losses of each task.

~\textbf{Face verification task} The face verification task  is trained by the class-wise triplet loss~\cite{ming2017simple} joint with the cross-entropy loss of softmax. The class-wise triplet loss acts as a regularization term of the cross-entropy loss. The loss function of face verification task $L_1$ is given by:      
\begin{equation}
\label{eq1}
L_1 = L_{s1} + \alpha{L_{c}}
\end{equation}
where $L_{s1}$ is the cross-entropy loss of task 1, $L_{c}$ is the class-wise triplet loss weighted by the $\alpha$. The cross-entropy loss $L_{s1}$ is given by:
\begin{equation}
L_{s1}= -\sum^{m}_{i=1}\sum^{k}_{j=1}1\{y_i=j\}log\frac{e^{z_j}}{\sum^k_{l=1}e^{z_l}}
\end{equation}
\label{eq:cross_entropy}

where $m$ is the size of the mini-batch, $k$ is the number of the identities in training dataset $D$, ${y_i}$ is the identity label of the input feature $x_i$ corresponding to the image $I_i \in D$, $z_j$ is output of the last fully-connected layer of BRANCH 1.

The class-wise triplet loss is a loss function aiming to simplify the triplet loss~\cite{schroff2015facenet}. Instead of minimizing the margin of an anchor $x_i$ to the positive example with the same class and to the negative example in all possible triplets, the class-wise triplet loss only computes the margin of the anchor $x_i$ to the intra-class center and to the inter-class centers. Then the class-wise triplet loss $L_c$ is given by: 
\begin{equation}
\begin{split}
L_{c} =\sum_{i=1}^m{\sum_{l=1,l\neq{y_i}}^{k}\max(d_{y_i,i}+\beta_0-d_{l,i}, 0)}
\end{split}
\end{equation}
where $m$ is the size of the mini-batch, $k$ is the number of the classes, i.e. the number of the identities, ${y_i}$ is the identity label of $x_i$, $d_{y_i,i}$ is the Euclidean distance of the anchor ${x_i}$ to the intra-class center $c_i$ and $d_{l,i}$ is the distance of the anchor $x_i$ to the inter-center center $c_l$:
\begin{equation}
\label{loss_batch}
\begin{split}
d_{y_i,i} = ||x_i-c_i||,\quad d_{l,i} = ||x_i-c_l||,  
\end{split}
\end{equation}
$\beta_0$ is a hyperparameter to set the minimum margin between positive distance $d_{y_i,i}$ and the negative distance $d_{l,i}$ .

~\textbf{Facial expression recognition task}
The loss function for facial expression recognition task  $L_2$ is a cross-entropy loss based on the softmax of the last fully connected layer of BRANCH 2. The equation of $L2$ is the same than $L_{s1}$ except the $k$ in $L_{2}$ is the number of the categories of the facial expressions rather than the number of identities in training dataset $D$. 

~\textbf{Dynamic multi-task learning}
The overall loss function for the proposed multi-task CNNs $L_3$ is the sum of the weighted $L_1$ and $L_2$, which is given by:

\begin{equation}
\label{eq_toatlloss}
L_3 = (1+w_1)L_{1} + w_2{L_{2}},
\end{equation}

where $w_1$, $w_2$ are the outputs of the softmax layer of the dynamic-weight-unit given by: 
\begin{equation}
\label{eq1_dynamicweights}
w1 = \frac{e^{z_1}}{\sum^2_{l=1}e^{z_l}}, \quad w2 = \frac{e^{z_2}}{\sum^2_{l=1}e^{z_l}}
\end{equation}
where the $z_l$ is the flat output of the last layer of the sharing hidden layers. Setting $1+w_1$ as the weight of face verification task is due to the training of face verification task which is more difficult than facial expression task. Once the weight of $w1$ varies to zero in the beginning of the training, the weighted loss of face verification will decay to zero in the overall loss function so that the face verification task will be no longer trained. Let $\mathbf{\epsilon}$ denotes the weights of the fully connected layer in dynamic-weights-unit, $\mathbf{W, b}$ denote the weights and bias of the convolutional filters and the weights of fully connected layers in the two branches, and $\mathbf{\gamma, \beta}$ denote the parameters of the batch normalization layers, the parameters are optimized to minimize the overall loss of the multi-task CNNs:

\begin{equation}
\label{tatolloss}
\mathop{\arg\min}_{\mathbf{W, b, \gamma, \beta, \epsilon}} (1+w_1)L_{1} + w_2{L_{2}}
\end{equation}

~\subsection{Training protocol}
Before the training of the proposed multi-task CNNs, a single-task network constituted of the sharing hidden layers and the BRANCH 1 is pretrained for face verification-task with large-scale dataset by loss function $L_1$. Then the training of the dynamic multi-task CNNs can handling on the pretrained model with the loss function $L_3$ Moreover, in order to compare the multi-task learning with the single-task learning, the BRANCH 2 is also trained independently by transferring the learning of the pretrained BRANCH 1 for facial expression recognition with loss function $L2$. Finally we obtain two models trained by the single-task learning for face verification (sharing layers + BRANCH 1) and facial expression recognition (sharing layers + BRANCH 2) respectively, and a dynamic multi-task CNNs trained by jointly multi-task learning.

\section{Experiments and analysis}
In this section we firstly evaluate the performance of the proposed dynamic multi-task model in comparison with the single-task learning models; and then analyzing the effectiveness of the automatic learning of the dynamic weights of tasks. Finally we evaluate the effectiveness of the dynamic multi-task FaceLiveNet+ for the holistic performance for face authentication. 

\subsection{Evaluation of the dynamic multi-task learning}

 Since the proposed multi-task networks performs the face verification task and the facial expression recognition task simultaneously, the datasets including both identity labels and facial expression labels are necessary to the training and the evaluation of the model. Normally the large-scale datasets used for training face recognition do not include the facial attributes such as facial expressions or only having some simple attributes such as smile, moustache, gender etc., as in Celeb-A~\cite{liu2015faceattributes}. Meanwhile some datasets used for training facial expression recognition task can  obtain the identity labels such as the OuluCASIA and CK+ used in this work as shown in Table~\ref{table1:expressiondatasets} (following the protocol in~\cite{liu2014learning} for extracting the images from the videos). However the size of these datasets are relatively small to train a deep CNNs-based face verification model from scratch. The other facial expression datasets such as FER2013~\cite{goodfellow2013challenges} collecting the images from internet in a wild condition has relative large size of images (more than 35k images), however it does not include the identity information. This is probably one of the reasons why the face verification joint with facial expression recognition multi-task learning has not been sufficiently studied. Therefore a face verification model with single-task learning is pretrained firstly in this study.
 
 \setlength\tabcolsep{4 pt}
\begin{table}
\centering
\small
\caption{The datasets used in the multi-task learning for face verification and facial expression recognition in this work. The labels of images is: ID (identification), Neutral (Ne), Anger (An), Disgust (Di), Fear (Fe), Happy (Ha), Sad (Sa), Surprise (Su), Contempt (Co).}
\label{table1:expressiondatasets}
\begin{tabular}{@{}l|ccccccccc@{}}
\hline
\multicolumn{1}{c}{} & ID & Ne & An & Di & Fe & Ha & Sa & Su & Co  \\
\hline
CK+  & 123 & 327 & 135 &  177 & 75  & 147  & 84  & 249  & 54  \\
OuluCASIA & 560& 560   & 240 & 240  &  240& 240  & 240  & 240  & -  \\
\hline

\end{tabular}
\end{table}

~\textbf{Pretrained face verification CNNs}  Prior to the multi-task learning, a single-task face verification deep CNNs with the structure mentioned in Section 4 is trained  based on the large scale dataset MSCeleb-1M. In both training and evaluation phase, the faces have been detected by the MTCNN~\cite{zhang2016joint} from the given raw images. The RMSprop with the mini-batches of 90 samples are applied for optimizing the parameters. The learning rate is started from 0.1, and decay by 10 at the 60K, 80K iterations respectively. The networks are initialized by Xavier~\cite{glorot2010understanding} and biases values are set to zero at beginning. The momentum coefficient is set to 0.99.  The dropout with the probability of 0.5 and the weight decay of 5e-5 are applied.  The weight of the class-wise triplet loss $\alpha$ is set to 1e-4, the margin $\beta_0$ is set to 10. ~\tableautorefname~\ref{tab:configurationAB_lfw} shows that the pretrained face verification model with single-task learning can achieve the state-of-art performance on the widely used datasets LFW and YTF. 

\begin{table}
\caption{\label{tab:configurationAB_lfw}The evaluation of the pretrained face verification CNNs with single-task learning on LFW and YTF datasets (accuracy\%).}
\begin{center}
\small
\begin{tabular}{|l|c|c|c|c|}
\hline
Method & Images & Nets & LFW & YTF.\\
\hline\hline
Fisher Faces~\cite{simonyan2013fisher}&-&-&${93.10}$&${83.8}$  \\\hline
DeepFace~\cite{taigman2014deepface} &4M&3&${97.35}$&${91.4}$  \\\hline
DeepID-2,3~\cite{sun2015deeply} &-&${200}$ &${99.47}$&${93.2}$  \\\hline
FaceNet~\cite{schroff2015facenet} &200M&1&${99.63}$&${95.1}$  \\\hline
VGGFace~\cite{simonyan2014very} &2.6M&1&${98.95}$&${91.6}$  \\\hline
Centerloss~\cite{wen2016discriminative} &0.7M&1&${99.28}$&${94.9}$  \\\hline
SphereFace~\cite{liu2017sphereface} &0.7M&1&${99.42}$&${95.0}$  \\\hline
\textbf{Pretrained (ours)} & 1.1M &1 & \textbf{99.42}& \textbf{95.0} \\
\hline
\end{tabular}
\end{center}

\end{table} 

~\textbf{ Multi-task learning for FaceLiveNet+}
The superiority of the proposed FaceLiveNet+ allows to perform face verification and facial expression simultaneously. Hence we will evaluate the performance of the proposed multi-task CNNs in both face verification and facial expression recognition.

\begin{figure}[t]
\begin{center}
   \includegraphics[width=1.0\linewidth]{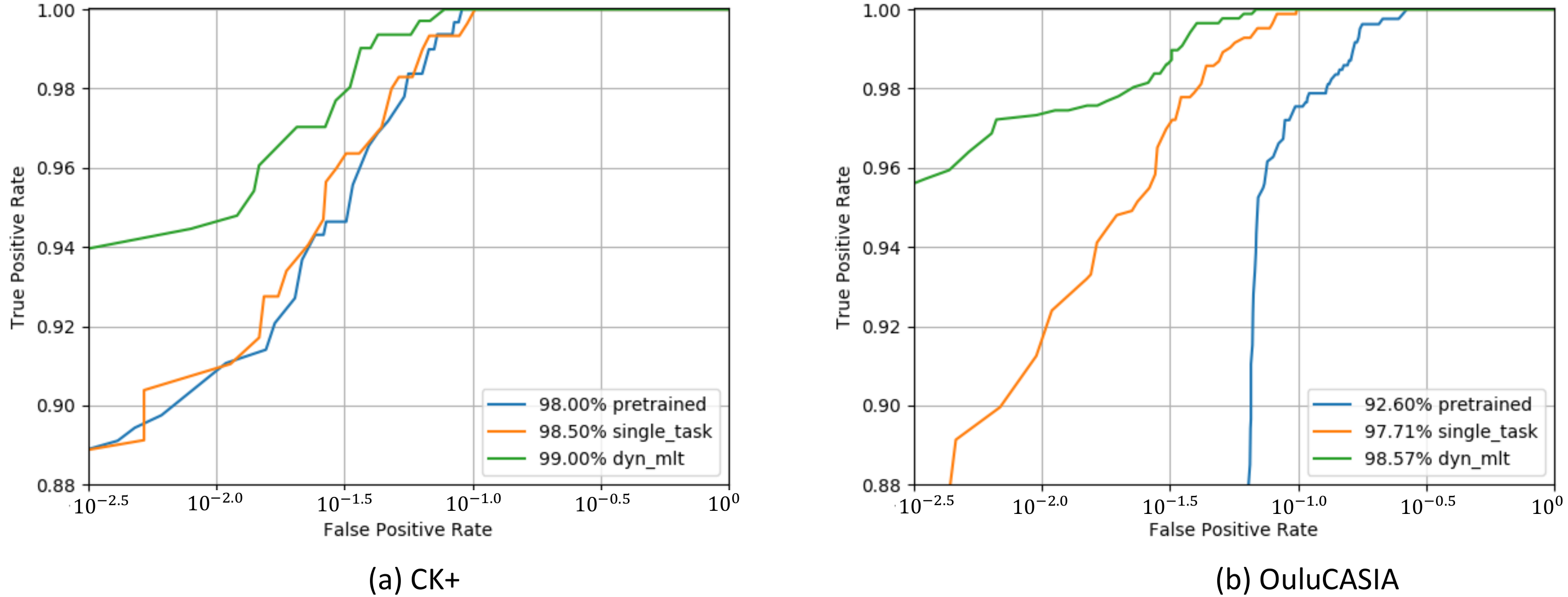}
\end{center}
   \caption{Comparison of the proposed dynamic multi-task model with single-task model and pretained CNNs on datasets (a) CK+  and (b) OuluCASIA for face verification.}
\label{fig:roc}
\end{figure}

\begin{figure}[t]
\begin{center}
   \includegraphics[width=1.0\linewidth]{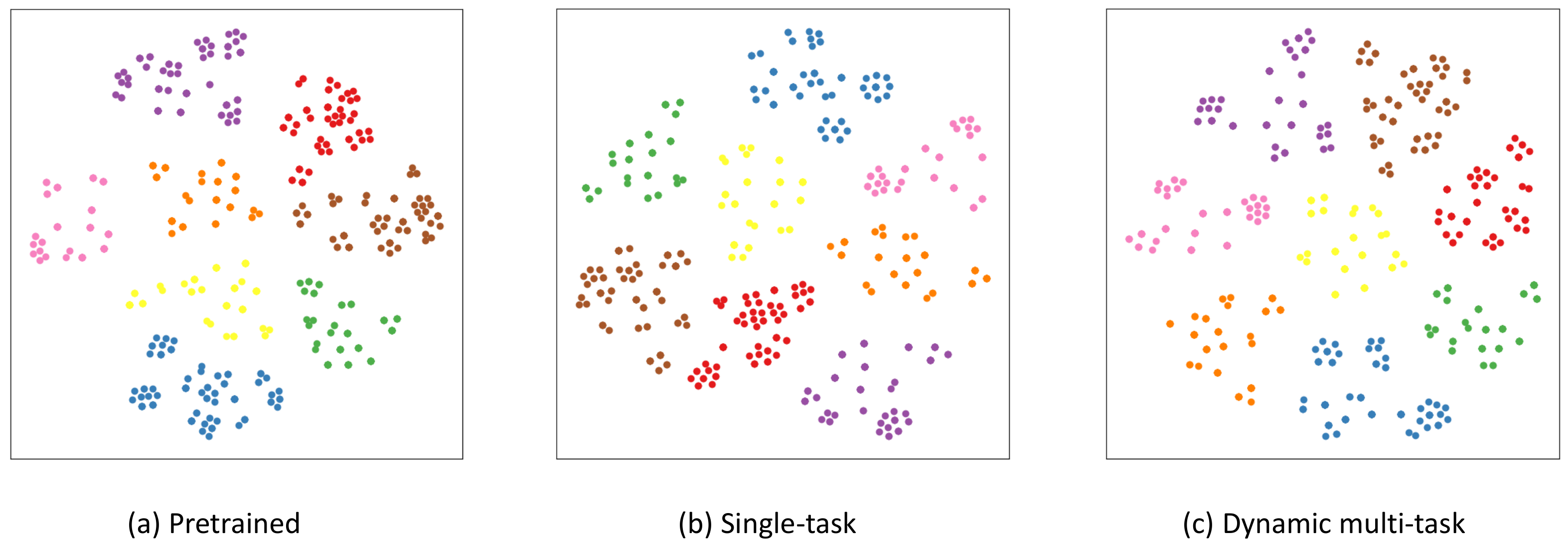}
\end{center}
   \caption{Visualization of the embedding feature space learned by the pretrained, single-task and dynamic multi-task models on dataset OuluCASIA by t-SNE. Each point is corresponding to an image of an individual. The different colors denote the different identities.}
\label{fig:tsne}
\end{figure}

~\textbf{a) face verification performance}
 Unlike the variations of the pose, the facial expression introduces  non-rigid distortions of the face which make difficult the face verification on images with facial expression variation. Table~\ref{tab:multitaskeval} shows the face verification on the  images with facial expressions. 
 As far we know, we are the first to evaluate the state-of-art face verification algorithms trained on general large-scale datasets with  facial expression dataset. 
 We can see that as well as our pretrained face verification CNNs, the performance of the state-of-art methods such as VGGFace, FaceNet and DeepID have degraded on the face images with facial expression, e.g. the accuracy of DeepID has decreased from 97.3\% to 91.7\% on CK+, VGGFace has decreased from 98.95\% to 92.2\% and even FaceNet has decreased slightly. This is quite probably resulted by the lack of the facial expression images in the general datasets for the training of the models. By fine-tuning our pretrained CNNs-based model with the facial expression datasets, the performance can obviously be improved (evaluating by the 10 folds cross-validation), e.g. the accuracy of our single-task fine-tuning model on OuluCASIA has improved from 92.6\% to 97.71\% of . Benefiting from the proposed dynamic multi-task model, the results can be further improved to obtain the best results both on the datasets CK+ and OuluCASIA. The dynamic multi-task model is also trained based on the pretrained CNNs as well as the single-task fine-tuning model. From the ROC curves shown by \figurename~\ref{fig:roc}, we can see that on CK+ the single-task fine-tuning model has not essentially improved the performance while the proposed multi-task model is evidently better than the other two models. By using t-SNE~\cite{maaten2008visualizing}, ~\figurename~\ref{fig:tsne} shows the learned embedding features space of the different models used for discriminating the identities on dataset OuluCASIA. Comparing to the distribution of the features learned by pretrained and single-task model, the multi-task model shows it can learned a better discriminable feature space.
 
 \begin{table}
\caption{\label{tab:multitaskeval}The evaluation of face verification on facial expressions datasets with different methods (accuracy\%).}
\begin{center}
\small
\begin{tabular}{|l|c|c|c|c|}
\hline
Method & Images & Nets & CK+ & OuluCASIA.\\
\hline\hline
DeepID-2,3 &-&${200}$ &${91.70}$&${96.50}$  \\\hline
FaceNet &200M&1&${98.00}$&${97.50}$  \\\hline
VGGFace &2.6M&1&${92.20}$&${93.5}$  \\\hline
\textbf{Pretrained (ours)}  & 1.1M &1 & \textbf{98.00}& \textbf{92.60} \\
\textbf{Single-task (ours)}  & 1.1M &1 & \textbf{98.50}& \textbf{97.71} \\
\textbf{Multi-task (ours)} & 0.46M &1.1 & \textbf{99.00}& \textbf{98.57}
\\
\hline
\end{tabular}
\end{center}

\end{table}

\begin{figure}[t]
\begin{center}
   \includegraphics[width=0.9
   \linewidth, angle =0]{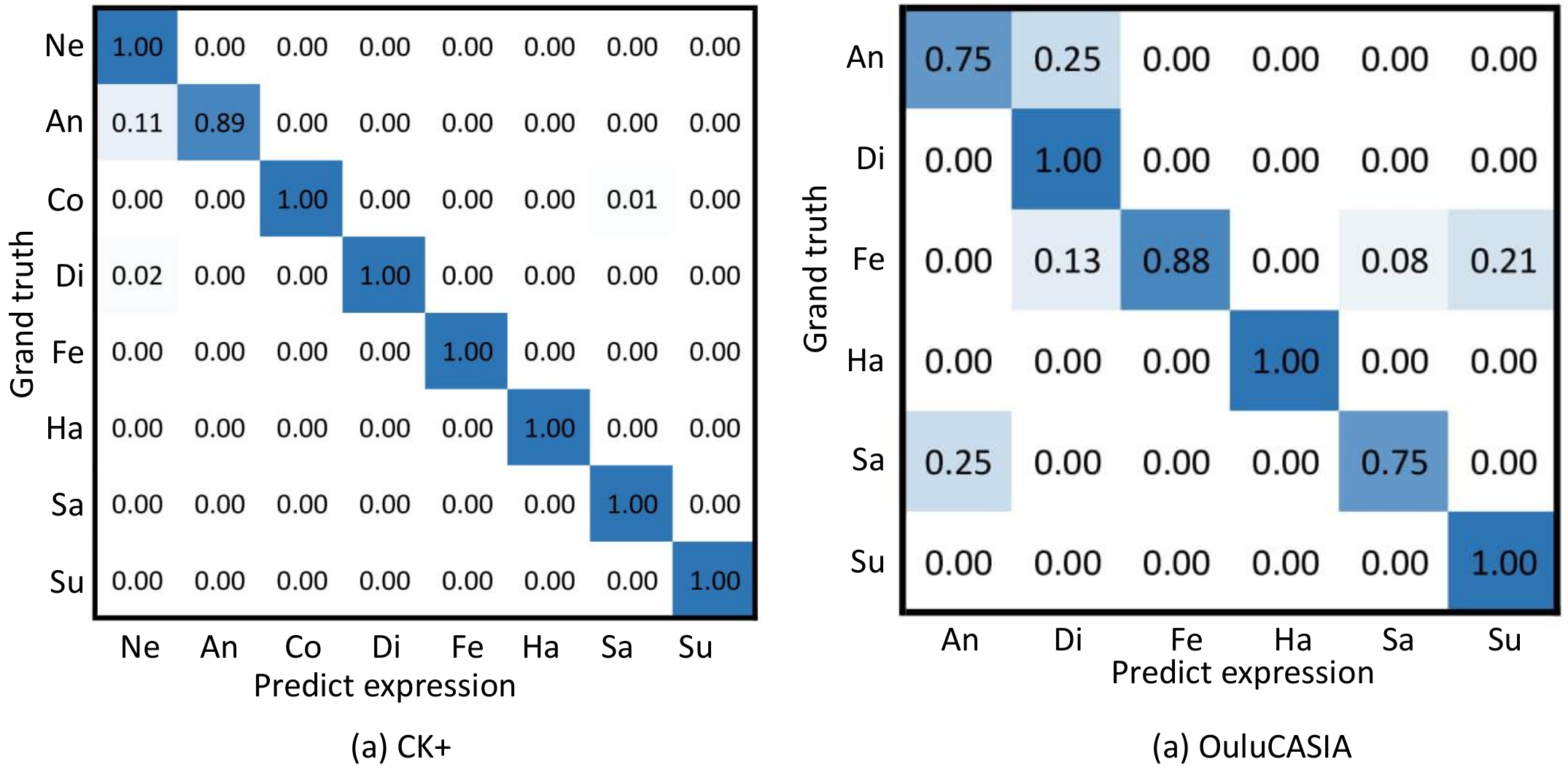}
\end{center}
   \caption{The confusion matrices of facial expression recognition of dynamic multi-task model on CK+ and OuluCASIA. The vertical axis is the ground truth label and the horizontal axis is the predicted expression. The darker color the higher accuracy.}
\label{fig:confusion}
\end{figure}

~\textbf{b) facial expression performance}
Table~\ref{tab:accexpression} compares the proposed multi-task networks for facial expression recognition with other methods on CK+ and OuluCASIA datasets. The single-task facial expression recognition model listed in the table is trained by transfer learning from the pretrained CNNs as mentioned in Section 4. Both the single-task and multi-task learning models have slightly outperform the state-of-art performance, however the multi-task model can further improve the performance comparing to the single-task model. ~\figurename~\ref{fig:confusion} shows the confusion matrices of facial expression recognition given by the dynamic multi-task model.

 \begin{table}
\begin{center}
\small
\caption{\label{tab:accexpression}The evaluation of proposed multi-task networks for facial expression recognition task on CK+ and OuluCASIA datasets.}
\begin{tabular}{cc}
\begin{minipage}[b]{0.22\textwidth}
    \begin{tabular}{|c|c|c|c|c|}
      \hline
      Method & Acc(\%) \\
      \hline\hline
      LBPSVM~\cite{feng2007facial}&${95.1}$   \\\hline
      Inception~\cite{mollahosseini2016going} &${93.2}$  \\\hline
      DTAGN~\cite{jung2015joint} &${97.3}$ \\\hline
      PPDN~\cite{zhao2016peak}&${97.3}$   \\\hline
      AUDN~\cite{liu2013aware} &${92.1}$   \\\hline
            
      \textbf{Single-task (ours)} &\textbf{98.21}   \\
      \textbf{Multi-task (ours)} &\textbf{99.10}   \\
      \hline
      \end{tabular}\\\\
      \centering (a) CK+

    \end{minipage}
    &\begin{minipage}[b]{0.21\textwidth}
		\begin{tabular}{|l|c|c|c|c|}
		\hline
      Method & Acc(\%) \\
      \hline\hline
      HOG3D~\cite{klaser2008spatio}&${70.63}$   \\\hline
      AdaLBP~\cite{zhao2011facial}&${73.54}$ \\\hline
      DTAGN~\cite{jung2015joint} &${81.46}$ \\\hline
      PPDN~\cite{zhao2016peak}&${84.59}$   \\\hline
      \textbf{Single-task (ours)} &\textbf{85.42}   \\
      \textbf{Multi-task (ours)} &\textbf{89.60}   \\\hline
      \end{tabular}\\\\
      \centering (b) OuluCASIA

    \end{minipage}

\end{tabular}
\end{center}

\end{table}

\begin{figure}[t]
\begin{center}
   \includegraphics[width=0.9
   \linewidth, angle =0]{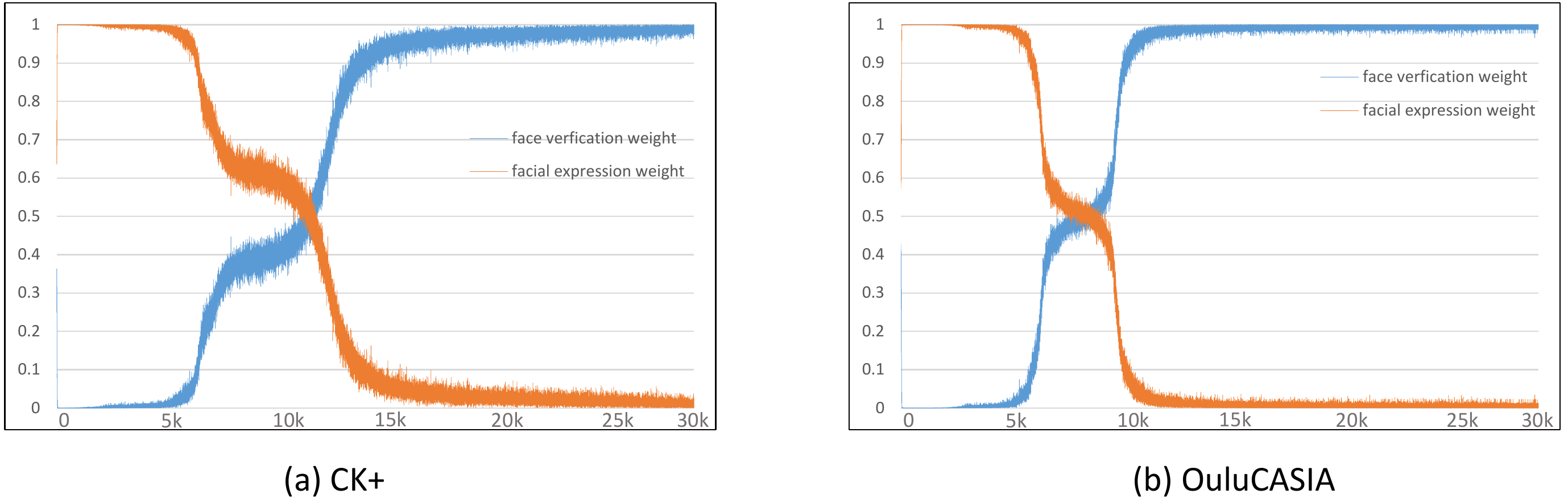}
\end{center}
   \caption{The dynamic weights of the tasks automatically learned by the networks. The blue one is the dynamic weight of the face verification task and the orange is the dynamic weight of the facial expression recognition task.}
\label{fig:dyn_weights}
\end{figure}

\begin{figure}[t]
\begin{center}
   \includegraphics[width=0.9
   \linewidth, angle =0]{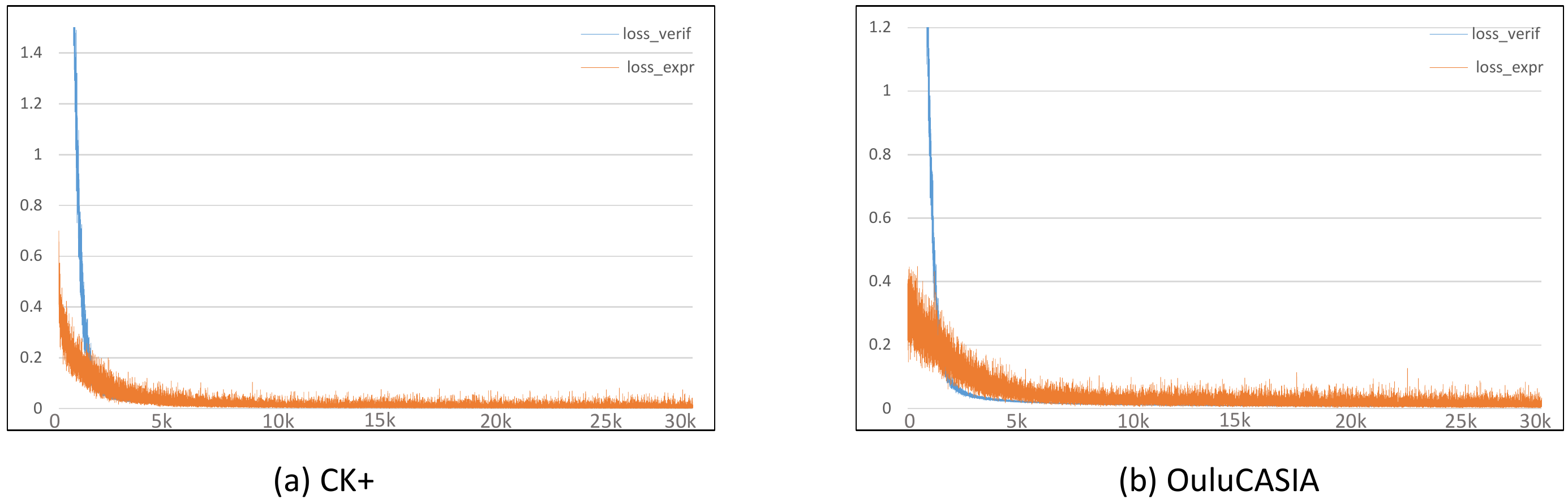}
\end{center}
   \caption{The loss of the two tasks of face verification and facial expression recognition during the training of the proposed dynamic multi-task model on the datasets CK+ and OuluCASIA.}
\label{fig:loss_ck_oulu}
\end{figure}

~\textbf{Dynamic weights Learning} Rather than the conventional multi-task method with fixed weights, the proposed dynamic multi-task model can automatically learn the weights of tasks to measure the importance of the tasks in real-time during the training processing. As shown in ~\figurename~\ref{fig:dyn_weights}, the weights of the tasks in the dynamic multi-tasks networks are automatically learned during the training of the networks. At the beginning, the parameters of the dynamic-weights-unit are initialized with the random value as well as other parts of the networks, thus the weights generated by the units are more or less 0.5 for each task. Very soon, the weight of the facial expression recognition task (the blue one) boost to 1 and the weight of face verification drop to zero. It means that the networks tries to focus on the training of the facial expression recognition task. Later, when the training of the face expression recognition task become saturated the network decreases the weight of this task and switches to train the other task by increasing its weight for augmenting its importance in the overall training. This point can be observed clearly by the variation of the two weights. The variation of the weights is coherent on both of two datasets which proves the effectiveness of the automatic learning of the weights of the tasks in our proposed multi-task learning framework. ~\figurename~\ref{fig:loss_ck_oulu} has shown the loss variation of the two tasks of face verification and facial expression recognition during the dynamic multi-task learning. One can note that the loss of face verification continues to decrease even when the weight of task decreases to zeros. Actually, the face verification task is weighted by $1+w_1$ as shown in equation~\ref{eq_toatlloss}, so even the dynamic weight $w_1$ has decreased to zero, the task of face verification can still keep in the overall optimization. Since the face verification task is normally much more difficult to be trained than the facial expression recognition task, if its dynamic weight drops to zero at the very beginning of the training phase the task will never to be trained without a constant variable weight in the overall loss function. As shown in Table~\ref{table1:expressiondatasets}, the dynamic multi-task learning has also be compared with the multi-task learning with fixed weights on dataset OuluCASIA. Except the setting of the weights of the tasks, the configurations of the two frameworks are exactly same. Three combinations of weights roughly represent the three relationship between the importance of the tasks, i.e. equal, more important and less important. The results show that the proposed dynamic multi-task learning is superior to the static multi-task learning with fixed weights.

\begin{table}
\caption{\label{tab:staticmulti}Comparison of the proposed dynamic multi-task learning with the static multi-task learning with the fixed weights (accuracy\%) on dataset OuluCASIA.}
\begin{center}
\small
\begin{tabular}{|l|c|c|c|c|}
\hline
Method & $w1$ & $w2$ & face\_verif & facial\_expr\\
\hline\hline
Static multi-task &1&1&${96.57}$&${88.2}$  \\
 &1&2&${96.00}$&${88.2}$  \\
 &1&0.5 &${97.14}$&${88.9}$  \\\hline
Dynamic multi-task &-&-&\textbf{98.57}&\textbf{89.6}  \\\hline


\end{tabular}
\end{center}

\end{table} 
\subsection{Holistic evaluation for face authentication}
In this section, we employ a holistic evaluation for face authentication with the proposed framework. Since so far there is no dataset or protocol that can be used directly for evaluating the face authentication system by fusing face verification and facial expression recognition with the challenge-response mechanism, we propose a simple protocol to evaluate the face authentication system. Images pairs extracted from the CK+ or OuluCASIA are used as the test samples for the evaluation: one image is used as the user input image and the other one is the reference image, meanwhile the user input image is also used to do the liveness detection based on face verification. Thus a pair of images with two labels, i.e. the label for face verification denoted as true or false, and the label of a facial expression as the system requirement. Only when the two conditions are satisfied, the face authentication is positive. A test example for the evaluation can be given by Table~\ref{tab:Faceauthen_example}. Finally 2K image pairs are extracted from CK+ and 5K image pairs are extracted from OuluCASIA as shown in Table~\ref{tab:dataauthen}. Specifically, in this work only 'Happy' and 'Surprise' are adopted as the required expressions for liveness detection since these two expressions are most universal and easy to recognize.  The results for a holistic evaluation for face authentication with our proposed dynamic multi-task framework is shown in Table~\ref{tab:globaleval}. The favorable results show that applying the proposed multi-task FaceLiveNet+ is relevant for face authentication.

\begin{table}
\caption{\label{tab:Faceauthen_example} A test example for evaluating face authentication.}
\begin{center}
\small
\begin{tabular}{|l|c|c|c|c|}
\hline
image1 & image2 & Same ID? & Required expression \\
\hline

\end{tabular}
\end{center}

\end{table}

\begin{table}
\caption{\label{tab:dataauthen} The positive and negative pairs in the evaluation datasets for face authentication extracted from CK+ and OuluCASIA.}
\small
\begin{center}
\begin{tabular}{lcccc}

&\begin{minipage}[b]{0.23\textwidth}
		\begin{tabular}{|l|c|c|c|c|}
		\hline
       & ID-True & ID-False \\
      \hline
      Ex-True &114 & 1062\\
      \hline
      Ex-False &494 & 429\\     
      \hline
      \end{tabular}\\\\
      \centering (a) CK+

    \end{minipage}
    &\begin{minipage}[b]{0.22\textwidth}
		\begin{tabular}{|l|c|c|c|c|}
		\hline
       & ID-True & ID-False \\
      \hline
      Ex-True &288 & 2016\\
      \hline
      Ex-False &1440 & 1440\\     
      \hline
      \end{tabular}\\\\
      \centering (b) OuluCASIA+

    \end{minipage}
\end{tabular}
\end{center}

\end{table} 

\begin{table}
\begin{center}
\caption{\label{tab:globaleval}The holistic evaluation of face authentication with the proposed dynamic multi-task framework. $Acc_{auth}$ is the accuracy of face authentication, $Acc_{verif}$ is the accuracy of face verification, $Acc_{expre}$ is the accuracy of facial expression recognition, $Acc_{live}$ is the accuracy of liveness detection} 
\small
\begin{tabular}{|l|c|c|c|c|}
\hline
 &$Acc_{auth}$&$Acc_{verif}$ &$Acc_{expre}$&$Acc_{live}$ \\
\hline
CK+ &${0.960}$& ${0.960}$ & ${1.000}$ &${1.000}$\\\hline
OuluCASIA &${0.986}$& ${0.986}$& ${0.932}$ &${1.000}$\\\hline
\end{tabular}
\end{center}

\end{table} 


\section{Conclusion}
In this work, We have proposed a holistic dynamic multi-task CNNs for face authentication namely FaceLiveNet+. Benefiting from the multi-task framework, the FaceLiveNet+ can perform face verification and facial expression recognition simultaneously. Rather than the conventional multi-task deep networks with the fixed weights of the tasks, a so-called dynamic-weights-unit is proposed to learn the weights of the tasks automatically in our proposed dynamic-task deep CNNs. Experimental results show that the proposed multi-task framework with the dynamic weights is superior to the deep networks trained by the single-task learning, especially for face verification with the facial expression images which is a problem that is not extensively studied. The experiments have also proved the effectiveness of the automatic learning of the weights of tasks, by which the multi-task networks can switch the training of the tasks by dynamically varying the weights of the tasks. The comparison with the static multi-task framework with the fixed weights shows the superiority of the the proposed multi-task framework with the dynamic weights of tasks. Finally, the holistic evaluations for face authentication with the proposed FaceLiveNet+ show the relevance for face authentication.

{\small
\bibliographystyle{ieee}
\bibliography{egbib_final}
}

\end{document}